\pdfoutput=1

\documentclass[11pt]{article}

\usepackage{acl}

\usepackage{times}
\usepackage{latexsym}
\usepackage[russian,english]{babel}
\usepackage[multiple]{footmisc}

\usepackage[T1]{fontenc}

\usepackage[utf8]{inputenc}

\usepackage{microtype}

\title{Exploring Cross-lingual Textual Style Transfer with \\ Large Multilingual Language Models}

  \author{Daniil Moskovskiy$^1$~~~~~~~ Daryna Dementieva$^{1,2}$~~~~~~~ Alexander Panchenko$^1$ \\
  $^1$Skolkovo Institute of Science and Technology, Russia\\
  $^2$Technical University of Munich, Germany\\
  \texttt{\href{mailto:daniil.moskovskiy@skoltech.ru}{\{daniil.moskovskiy,daryna.dementieva,a.panchenko\}@skoltech.ru}} }

\date{}

\begin{document}
\maketitle
\begin{abstract}
Detoxification is a task of generating text in polite style while preserving meaning and fluency of the original toxic text. Existing detoxification methods are designed to work in one exact language. This work investigates  multilingual and cross-lingual detoxification and the behavior of large multilingual models like in this setting. Unlike previous works we aim to make large language models able to perform detoxification without direct fine-tuning in given language. Experiments show that multilingual models are capable of performing multilingual style transfer. However, models are not able to perform cross-lingual detoxification and direct fine-tuning on exact language is inevitable. 
\end{abstract}

\section{Introduction}

The task of Textual Style Transfer (Textual Style Transfer) can be viewed as a task where certain properties of text are being modified while rest retain the same\footnote{Hereinafter the data-driven definition of style is used. Therefore, we call style a characteristic of given dataset that differs from a general dataset \cite{jin2021deep}.}. In this work we focus on detoxification textual style transfer \cite{santos2018fighting,dementieva2021methods}. It can be formulated as follows: given two text corpora $D^X=\{x_1, x_2, \dots x_n\}$ and $D^Y=\{y_1, y_2, \dots, y_n\}$, where $X$, $Y$ - are two sets of all possible text in styles $s^X$, $s^Y$ respectively, we want to build a model $f_{\theta} : X \rightarrow Y$, such that the probability $p(y_{gen}|x, s^X, s^Y)$ of transferring the style $s^X$ of given text $x$ (by generation $y_{gen}$) to the style $s^Y$ is maximized (where $s^X$ and $s^Y$ are toxic and non-toxic styles respectively).

Some examples of detoxification presented in Table \ref{tab:detoxification_example}.

Textual style transfer gained a lot of attention with a rise of deep learning-based NLP methods. Given that, Textual Style Transfer has now a lot of specific subtasks ranging from formality style transfer \cite{rao-tetreault-2018-dear, yao2021improving} and simplification of domain-specific texts \cite{devaraj2021paragraphlevel, maddela2021controllable} to emotion modification \cite{sharma2021facilitating} and detoxification (debiasing) \cite{li2020stylecontent, dementieva2021methods}. 

\begin{table*}[]
\centering
\begin{tabular}{c|c}
\hline
\textbf{Source text}         & \textbf{Target text}  \\ \hline
What the f*ck is your problem?  & What is your problem?     \\
This whole article is bullshit. & This article is not good. \\
Yeah, this clowns gonna make alberta great again! & Yeah, this gonna make Alberta great again \\ \hline
\end{tabular}
\caption{Examples of desired detoxification results.}
\label{tab:detoxification_example}
\end{table*}

There exist a variety of Textual Style Transfer methods: from totally \textbf{supervised} methods \cite{Wang2019HarnessingPN,zhang2020parallel,dementieva2021methods} which require a parallel text corpus for training to \textbf{unsupervised} \cite{shen2017style,wang2019controllable,xu2021vae} that are designed to work without any parallel data. The latter sub-field of research is more popular nowadays due to the scarcity of parallel text data for Textual Style Transfer. On the other hand, if we address Textual Style Transfer task as a Machine Translation task we get a significant performance boost \cite{prabhumoye2018style}. 

The task of detoxification, in which we focus in this work, is relatively new. First work on detoxification was a sequence-to-sequence collaborative classifier, attention and the cycle consistency loss \cite{nogueira-dos-santos-etal-2018-fighting}. A recent work by \cite{laugier2021civil} introduces self-supervised model based on T$5$ model \cite{raffel2020exploring}  with a denoising and cyclic auto-encoder loss. 

Both these methods are unsupervised which is an advantage but it comes from the major current problem of the textual style transfer. There is a lack of parallel data for Textual Style Transfer since there exist only few parallel datasets for English \cite{rao-tetreault-2018-dear} and some other languages \cite{briakou2021xformal}. When it comes to detoxification there are only two parallel detoxification corpora available now and they both appeared only last year \cite{dementieva2021crowdsourcing}. Most state-of-the-art methods rely on large amounts of text data which is often available for some well-researched languages like English but lacking for other languages almost entirely. Therefore, it is important to study whether cross-lingual (or at least multilingual) detoxification is possible.



Multilingual language models such as mBART \cite{liu2020multilingual}, mT5 \cite{xue2021mt5} have recently become available. This work explores the possibility of multilingual and cross-lingual textual style transfer (Textual Style Transfer) using such large multilingual language models. We test the hypothesis that modern large text-to-text models are able to generalize ability of style transfer across languages. 

Our contributions can be summarized as follows\footnote{All code is available online: \url{https://github.com/skoltech-nlp/multilingual_detox}}:
\begin{enumerate}
    \item We introduce a novel study of multilingual textual style transfer and conduct experiments with several multilingual language models and evaluate their performance. 
    \item We conduct cross-lingual Textual Style Transfer experiments to investigate whether multilingual language models are able to perform Textual Style Transfer without fine-tuning on a specific language.
\end{enumerate}

\section{Methodology}

We formulate the task of \textbf{supervised} Textual Style Transfer as a sequence-to-sequence NMT task and fine-tune multilingual language models to translate from "toxic" to "polite" language.

\subsection{Datasets}

In this work we use two datasets for Russian and English languages. Aggregated information about datasets could be found in Table \ref{table:kysymys}, examples from datasets can be found in \ref{sec:appendix_1} and \ref{sec:appendix_2}.

\begin{table}[h!]
\centering
\label{tab:table_1}
\begin{tabular}{l| l l l}
\hline
\textbf{Language}        & \textbf{Train}    & \textbf{Dev}    & \textbf{Test}   \\ \hline
English & $18 777$ & $988$  & $671$  \\ 
Russian & $5058$   & $1000$ & $1000$ \\
\hline
\end{tabular}
\caption{Aggregated datasets statistics.}
\label{table:kysymys}
\end{table}

\paragraph{Russian data} We use detoxification dataset\footnote{\url{https://github.com/skoltech-nlp/russe_detox_2022}} which consists of $5 058$ training sentences, $1 000$ validation sentences and $1 000$ test sentences.

\paragraph{English data} We use ParaDetox \cite{dementieva2021crowdsourcing} dataset. It consists of $19766$ \textit{toxic} sentences and their \textit{polite} paraphrases. This data is split into training and validation as $95\%$ for training and $5\%$ for validation. For testing we use a set of $671$ toxic sentences. 

\subsection{Experimental Setup}

We perform a series of experiments on detoxification using parallel data for English and Russian. We train models in two different setups: \textbf{multilingual} and \textbf{cross-lingual}. 

\paragraph{Multilingual setup} In this setup we train models on data containing both English and Russian texts
and then compare their performance with baselines trained on these languages solely.

\paragraph{Cross-lingual setup} In cross-lingual setup we test the hypothesis that models are able to perform detoxification without explicit fine-tuning on exact language. We fine-tune models on English and Russian separately and then test their performance.   

\subsection{Models}

Scaling language models to many languages has become an emerging topic of interest recently \cite{devlin2019bert, tan2019multilingual, lample2019crosslingual, conneau2020unsupervised}. We adopt several multilingual models to textual style transfer in our work. 


\paragraph{Baselines} We use two detoxification methods as baselines in this work - \textbf{Delete} method which simply deletes toxic words in the sentence according to the vocabulary of toxic words and \textbf{CondBERT}. The latter approach works in usual masked-LM setup by masking toxic words and replacing them with non-toxic ones. This approach was first proposed by \cite{condbert} as a data augmentation method and then adopted to detoxification by \cite{dale2021text}.  

\paragraph{mT5} mT5 \cite{xue2021mt5} is a multilingual version of T5 \cite{raffel2020exploring} - a text-to-text transformer model, which was trained on many downstream tasks. mT5 replicates T5 training but now it is trained on more than $100$ languages.

\paragraph{mBART} mBART \cite{liu2020multilingual} is a multilingual variation of BART  \cite{lewis2019bart} - denoising autoencoder built with a sequence-to-sequence model. mBART is trained on monolingual corpora across many languages. We adopt mBART in sequence-to-sequence detoxification task via fine-tuning on parallel detoxification dataset.  



\subsection{Evaluation metrics}

Unlike other NLP tasks, one metric is not enough to benchmark the quality of style transfer. The ideal Textual Style Transfer model output should \textit{preserve the original content} of the text, \textit{change the style} of the original text to target and the generated text also should \textit{be grammatically correct}. We follow \newcite{dale2021text} approach in Textual Style Transfer evaluation.


\subsubsection{Content Preservation}

\paragraph{Russian} Content preservation score (\textbf{SIM}) is evaluated as a cosine similarity of LaBSE \cite{feng2020languageagnostic} sentence embeddings. The model is slightly different from the original one, only English and Russian embeddings are left. 

\paragraph{English} Similarity (\textbf{SIM}) between the embedding of the original sentence and the generated one is calculated using the model presented by \newcite{wieting2019beyond}. Being is trained on paraphrase pairs extracted from ParaNMT corpus \cite{wieting-gimpel-2018-paranmt}, this model's training objective is to select embeddings such that the similarity of embeddings of paraphrases is higher than the similarity between sentences that are not paraphrases.

\subsubsection{Grammatic and language quality (fluency)}

\paragraph{Russian} We measure fluency (\textbf{FL}) with a BERT-based classifier \cite{devlin2019bert}
trained to distinguish real texts from corrupted ones. The model was trained on Russian texts and their corrupted (random word replacement, word deletion and insertion, word shuffling etc.) versions. Fluency is calculated as a difference between the probabilities of being corrupted for source and target sentences. The logic behind using difference is that we ensure that the generated sentence is not worse than the original one in terms of fluency.

\paragraph{English} We measure fluency (\textbf{FL}) as a percentage of fluent sentences evaluated by the RoBERTa-based\footnote{\url{https://huggingface.co/roberta-large}} \cite{liu2019roberta} classifier of linguistic acceptability trained on CoLA \cite{warstadt2019neural} dataset.

\subsubsection{Style transfer accuracy}

\paragraph{Russian} Style transfer accuracy (\textbf{STA}) is evaluated with a BERT-based \cite{devlin2019bert} toxicity classifier\footnote{\url{https://huggingface.co/SkolkovoInstitute/russian_toxicity_classifier}} fine-tuned from RuBERT Conversational. This classifier was additionally trained on Russian Language Toxic Comments dataset collected from \url{2ch.hk} and Toxic Russian Comments dataset collected from \url{ok.ru}. 

\paragraph{English} Style transfer accuracy (\textbf{STA}) is calculated with a style classifier - RoBERTa-based \cite{liu2019roberta} model trained on the union of three Jigsaw datasets \cite{jigsaw_toxic}. The sentence is considered toxic when the classifier confidence is above $0.8$. The classifier reaches the AUC-ROC of $0.98$ and F$_1$-score of $0.76$.


\subsubsection{Joint metric}

Aforementioned metrics must be properly combined to get one \textit{Joint} metric to evaluate Textual Style Transfer. We follow \newcite{krishna-etal-2020-reformulating} and calculate \textbf{J} as an average of products    of sentence-level \textit{fluency}, \textit{style transfer accuracy}, and \textit{content
preservation}: 
\begin{equation}
    \textbf{J} = \frac{1}{n}\sum\limits_{i=1}^{n}\textbf{STA}(x_i) \cdot \textbf{SIM}(x_i) \cdot \textbf{FL}(x_i) 
\end{equation}


\begin{table*}[!htbp]
\footnotesize
\centering
\begin{tabular}{lllllllll}
\hline
\multicolumn{1}{|l|}{} &
  \multicolumn{1}{c|}{\textbf{STA}$\uparrow$} &
  \multicolumn{1}{c|}{\textbf{SIM}$\uparrow$} &
  \multicolumn{1}{c|}{\textbf{FL}$\uparrow$} &
  \multicolumn{1}{c|}{\textbf{J}$\uparrow$} &
  \multicolumn{1}{c|}{\textbf{STA}$\uparrow$} &
  \multicolumn{1}{c|}{\textbf{SIM}$\uparrow$} &
  \multicolumn{1}{c|}{\textbf{FL}$\uparrow$} &
  \multicolumn{1}{c|}{\textbf{J}$\uparrow$} \\ \hline
\multicolumn{1}{|c|}{} &
  \multicolumn{4}{c|}{\textbf{Russian}} &
  \multicolumn{4}{c|}{\textbf{English}} \\ \hline  
\multicolumn{1}{|c|}{} &
  \multicolumn{8}{c|}{\it Baselines} \\ \hline
\multicolumn{1}{|l|}{Delete} &
  \multicolumn{1}{l|}{0.532} &
  \multicolumn{1}{l|}{0.875} &
  \multicolumn{1}{l|}{0.834} &
  \multicolumn{1}{l|}{0.364} &
  \multicolumn{1}{l|}{0.810} &
  \multicolumn{1}{l|}{0.930} &
  \multicolumn{1}{l|}{0.640} &
  \multicolumn{1}{l|}{0.460} \\ 
\multicolumn{1}{|l|}{condBERT~\cite{dale2021text} }  &
  \multicolumn{1}{l|}{{0.819}} &
  \multicolumn{1}{l|}{0.778} &
  \multicolumn{1}{l|}{0.744} &
  \multicolumn{1}{l|}{0.422} &
  \multicolumn{1}{l|}{\textbf{0.980}} &
  \multicolumn{1}{l|}{0.770} &
  \multicolumn{1}{l|}{0.820} &
  \multicolumn{1}{l|}{0.620} \\ \hline
\multicolumn{1}{|c|}{} &
  \multicolumn{8}{c|}{\it  Multilingual Setup} \\ \hline
\multicolumn{1}{|l|}{mT$5$ base} &
      \multicolumn{1}{l|}{0.772} &
  \multicolumn{1}{l|}{0.676} &
  \multicolumn{1}{l|}{0.795} &
  \multicolumn{1}{l|}{0.430} &
  \multicolumn{1}{l|}{0.833} &
  \multicolumn{1}{l|}{0.826} &
  \multicolumn{1}{l|}{0.830} &
  \multicolumn{1}{l|}{0.556} \\
\multicolumn{1}{|l|}{mT$5$ small} &
  \multicolumn{1}{l|}{0.745} &
  \multicolumn{1}{l|}{0.705} &
  \multicolumn{1}{l|}{0.794} &
  \multicolumn{1}{l|}{0.428} &
  \multicolumn{1}{l|}{0.826} &
  \multicolumn{1}{l|}{0.841} &
  \multicolumn{1}{l|}{0.763} &
  \multicolumn{1}{l|}{0.513} \\
\multicolumn{1}{|l|}{mT$5$ base$^{*}$} &
  \multicolumn{1}{l|}{{0.773}} &
  \multicolumn{1}{l|}{0.676} &
  \multicolumn{1}{l|}{0.795} &
  \multicolumn{1}{l|}{0.430} &
  \multicolumn{1}{l|}{0.893} &
  \multicolumn{1}{l|}{0.787} &
  \multicolumn{1}{l|}{\textbf{0.942}} &
  \multicolumn{1}{l|}{\textbf{0.657}} \\
\multicolumn{1}{|l|}{mBART $5000$} &
  \multicolumn{1}{l|}{0.685} &
  \multicolumn{1}{l|}{\textbf{0.778}} &
  \multicolumn{1}{l|}{\textbf{0.841}} &
  \multicolumn{1}{l|}{\textbf{0.449}} &
  \multicolumn{1}{l|}{0.887} &
  \multicolumn{1}{l|}{\textbf{0.889}} &
  \multicolumn{1}{l|}{0.866} &
  \multicolumn{1}{l|}{0.640} \\ \hline
\multicolumn{1}{|c|}{} &
  \multicolumn{8}{c|}{\it  Cross-lingual Setup} \\ \hline
\multicolumn{1}{|l|}{mT$5$ base ENG} &
  \multicolumn{1}{l|}{\textcolor{gray}{0.838}} &
  \multicolumn{1}{l|}{\textcolor{gray}{0.276}} &
  \multicolumn{1}{l|}{\textcolor{gray}{0.506}} &
  \multicolumn{1}{l|}{\textcolor{gray}{0.115}} &
  \multicolumn{1}{l|}{0.860} &
  \multicolumn{1}{l|}{0.834} &
  \multicolumn{1}{l|}{0.833} &
  \multicolumn{1}{l|}{0.587} \\
\multicolumn{1}{|l|}{mT$5$ base RUS} &
  \multicolumn{1}{l|}{0.676} &
  \multicolumn{1}{l|}{0.794} &
  \multicolumn{1}{l|}{0.846} &
  \multicolumn{1}{l|}{0.454} &
  \multicolumn{1}{l|}{\textcolor{gray}{0.906}} &
  \multicolumn{1}{l|}{\textcolor{gray}{0.365}} &
  \multicolumn{1}{l|}{\textcolor{gray}{0.696}} &
  \multicolumn{1}{l|}{\textcolor{gray}{0.171}} \\
\multicolumn{1}{|l|}{mT$5$ small ENG} &
  \multicolumn{1}{l|}{\textcolor{gray}{0.805}} &
  \multicolumn{1}{l|}{\textcolor{gray}{0.225}} &
  \multicolumn{1}{l|}{\textcolor{gray}{0.430}} &
  \multicolumn{1}{l|}{\textcolor{gray}{0.077}} &
  \multicolumn{1}{l|}{0.844} &
  \multicolumn{1}{l|}{0.858} &
  \multicolumn{1}{l|}{0.826} &
  \multicolumn{1}{l|}{0.591} \\
\multicolumn{1}{|l|}{mT$5$ small RUS} &
  \multicolumn{1}{l|}{0.559} &
  \multicolumn{1}{l|}{0.822} &
  \multicolumn{1}{l|}{0.817} &
  \multicolumn{1}{l|}{0.363} &
  \multicolumn{1}{l|}{\textcolor{gray}{0.776}} &
  \multicolumn{1}{l|}{\textcolor{gray}{0.521}} &
  \multicolumn{1}{l|}{\textcolor{gray}{0.535}} &
  \multicolumn{1}{l|}{\textcolor{gray}{0.169}} \\
\multicolumn{1}{|l|}{mBART $3000$ ENG} &
  \multicolumn{1}{l|}{\textcolor{gray}{0.923}} &
  \multicolumn{1}{l|}{\textcolor{gray}{0.395}} &
  \multicolumn{1}{l|}{\textcolor{gray}{0.552}} &
  \multicolumn{1}{l|}{\textcolor{gray}{0.202}} &
  \multicolumn{1}{l|}{0.842} &
  \multicolumn{1}{l|}{0.856} &
  \multicolumn{1}{l|}{0.876} &
  \multicolumn{1}{l|}{0.617} \\
\multicolumn{1}{|l|}{mBART $3000$ RUS} &
  \multicolumn{1}{l|}{0.699} &
  \multicolumn{1}{l|}{0.778} &
  \multicolumn{1}{l|}{0.858} &
  \multicolumn{1}{l|}{0.475} &
  \multicolumn{1}{l|}{\textcolor{gray}{0.547}} &
  \multicolumn{1}{l|}{\textcolor{gray}{0.778}} &
  \multicolumn{1}{l|}{\textcolor{gray}{0.888}} &
  \multicolumn{1}{l|}{\textcolor{gray}{0.299}} \\
\multicolumn{1}{|l|}{mBART $5000$ ENG} &
  \multicolumn{1}{l|}{\textcolor{gray}{0.900}} &
  \multicolumn{1}{l|}{\textcolor{gray}{0.299}} &
  \multicolumn{1}{l|}{\textcolor{gray}{0.591}} &
  \multicolumn{1}{l|}{\textcolor{gray}{0.160}} &
  \multicolumn{1}{l|}{0.857} &
  \multicolumn{1}{l|}{0.840} &
  \multicolumn{1}{l|}{0.873} &
  \multicolumn{1}{l|}{0.616} \\
\multicolumn{1}{|l|}{mBART $5000$ RUS} &
  \multicolumn{1}{l|}{0.724} &
  \multicolumn{1}{l|}{0.746} &
  \multicolumn{1}{l|}{0.827} &
  \multicolumn{1}{l|}{0.457} &
  \multicolumn{1}{l|}{\textcolor{gray}{0.806}} &
  \multicolumn{1}{l|}{\textcolor{gray}{0.484}} &
  \multicolumn{1}{l|}{\textcolor{gray}{0.864}} &
  \multicolumn{1}{l|}{\textcolor{gray}{0.242}} \\ \hline

\multicolumn{1}{|c|}{} &
\multicolumn{8}{c|}{\it Backtranslation Setup} \\ \hline

\multicolumn{1}{|l|}{mBART $5000$ (Google)}  &
  \multicolumn{1}{l|}{0.675} &
  \multicolumn{1}{l|}{0.669} &
  \multicolumn{1}{l|}{0.634} &
  \multicolumn{1}{l|}{0.284} &
 \multicolumn{1}{l|}{0.678} &
 \multicolumn{1}{l|}{0.762} &
 \multicolumn{1}{l|}{0.568} &
\multicolumn{1}{l|}{0.284} \\ 
  
\multicolumn{1}{|l|}{mBART $5000$ (FSMT)} &
\multicolumn{1}{l|}{0.737} &
\multicolumn{1}{l|}{0.633} &
\multicolumn{1}{l|}{0.731} &
\multicolumn{1}{l|}{0.348} &
 \multicolumn{1}{l|}{0.744} &
 \multicolumn{1}{l|}{0.746} &
 \multicolumn{1}{l|}{0.893} &
 \multicolumn{1}{l|}{0.415} \\ \hline
\end{tabular}
\caption{Evaluation of TST models.
Numbers in \textbf{bold} indicate the best results. $\uparrow$ describes the higher the better metric. Results of unsuccessful TST depicted as \textcolor{gray}{gray}. ENG and RUS depicts the data model have been trained on. mT$5$ base$^{*}$ was trained on all English and Russian data available (datasets were not equalized). Last row depicts backtranslation workaround for cross-lingual detoxification. We include only the best result for brevity.} 
\label{tab:table_2}
\end{table*}

\subsection{Training}

There is a variety of versions of large multilingual models available. In this work we use small and base versions of mT5\footnote{\url{https://huggingface.co/google/mt5-base}}\footnote{\url{https://huggingface.co/google/mt5-large}} \cite{xue2021mt5} and large version of mBART\footnote{\url{https://huggingface.co/facebook/mbart-large-50-many-to-many-mmt}} \cite{liu2020multilingual}.

\paragraph{Multilingual training}

In multilingual training setup we fine-tune models using both English and Russian data. We use Adam \cite{kingma2017adam} optimizer for fine-tuning with different learning rates ranging from $1 \cdot 10^{-3}$ to $5 \cdot 10^{-5}$ with linear learning rate scheduling. We also test different number of warmup steps from $0$ to $1000$. We equalize Russian and English data for training and use $10000$ toxic sentences and their polite paraphrases for multilingual training in total. We train mT5 models for $40$ thousand  iterations\footnote{According to \cite{xue2021mt5} mT5 was not fine-tuned on downstream tasks as the original T5 model. Therefore, model requires more fine-tuning iterations for Textual Style Transfer.} with a batch size of $8$. We fine-tune mBART \cite{liu2020multilingual} for $1000$, $3000$, $5000$ and $10000$ iterations with batch size of $8$. 

\paragraph{Cross-lingual training}

In cross-lingual training setup we fine-tune models using only one dataset, e.g.: we fine-tune model on English data and check performance on both English and Russian data. Fine-tuning procedure was left the same: $40000$ iterations for mT5 models and $1000$, $3000$, $5000$ and $10000$ iterations for the mBART. 

\textbf{Back-translation approach} to cross-lingual style transfer proved to work substantially better than the zero-shot setup discussed above. Nevertheless, both Google and FSMT did not yield scores comparable to monolingual setup. Besides, surprisingly Google yielded worse results than FSMT.

\section{Results \& Discussion}

Table \ref{tab:table_2} shows the best scores of both multilingual and cross-lingual experiments. In multilingual setup mBART performs better than baselines and mT$5$ for both English and Russian. Note that the table shows only the best results of the models. It is also notable that for mT$5$ increased training size for English data provides better metrics for English while keeping metrics for Russian almost the same. We also depict some of the generated detoxified sentences in the Table \ref{tab:table_2} in the part \ref{sec:appendix_b} of Appendix.

As for cross-lingual style transfer, results are negative. None of the models have coped with the task of cross-lingual Textual Style Transfer. That means that models produce the same or almost the same sentences for the language on which they were not fine-tuned so that toxicity is not eliminated. We provide only some scores here in the Table \ref{tab:results} for reference. 

Despite the fact that our hypothesis about the possibility of cross-language detoxification was not confirmed, the presence of multilingual models pre-trained in many languages gives every reason to believe that even with a small amount of parallel data, training models for detoxification is possible.

A recent work by \cite{multilingual_tst} shows that cross-lingual formality Textual Style Transfer is possible. \newcite{multilingual_tst} achieve this on XFORMAL dataset \cite{briakou2021xformal} by adding language-specific adapters in the vanilla mBART architecture \cite{liu2020multilingual} - two feed-forward layers with residual connection and layer normalization \cite{bapna-firat-2019-simple, DBLP:conf/icml/HoulsbyGJMLGAG19}. 

We follow the original training procedure described by \newcite{multilingual_tst} by training adapters for English and Russian separately on $5$ million sentences from News Crawl dataset\footnote{\url{https://data.statmt.org/news-crawl/}}. We use batch size of $16$ and $200$ thousand training iterations. We also then train cross-attentions on our parallel detoxifcation data in the same way. However, models tend to duplicate input text without any detoxification. 
Thus, while the exact same original setup did not work for detoxification, more parameter search and optimization could lead to more acceptable results and we consider the approach by \newcite{multilingual_tst} as a promising direction of a future work on multilingual and cross-lingual detoxification. 



\section{Conclusion}

In this work we have tested the hypothesis that multilingual language models are capable of performing cross-lingual and multilingual detoxification. In the multilingual setup we experimentally show that reformulating detoxification (Textual Style Transfer) as a NMT task boosts performance of the models given enough parallel data for training. We beat simple (Delete method) and more strong (condBERT) baselines in a number of metrics. Based on our experiments, we can assume that it is possible to fine-tune multilingual models in any of the $100$ languages in which they were originally trained. This opens up great opportunities for detoxification in unpopular languages.

However, our hypothesis that multilingual language models are capable of cross-lingual detoxification was proven to be false. We suggest that the reason for this is not a lack of data, but the model's inability to capture the pattern between toxic and non-toxic text and transfer it to another language by itself. This means that the problem of cross-lingual textual style transfer is still open and needs more investigation.





\section*{Acknowledgements}

This work was supported by MTS-Skoltech laboratory on AI.

\bibliography{anthology, custom}
\bibliographystyle{acl_natbib}

\newpage
\appendix
\onecolumn
\section{Data}
\subsection{English Dataset}\label{sec:appendix_1}

Table \ref{tab:ParaDetox} shows examples of sentence pairs from ParaDatex parallel detoxification corpora. There are several polite paraphrases for each toxic sentence in this dataset \cite{dementieva2021crowdsourcing}, this is a consequence of the way these parallel data are collected. Leaving only one paraphrase for one source sentence we could get $6000$ unique pairs of toxic sentences and their polite paraphrases. However, in this work we use data as is. 
\medskip

\begin{table*}[h!]
\centering
\begin{tabular}{l|l}
\hline
\color{red} Original & \color{red} my computer is broken and my phone too!! wtf is this devil sh*t???\\ \hline
Detoxed &  My computer is broken and my phone too! So disappointed!\\
& My computer is broken and my phone too, what is this?\\
& Both my computer and phone are broken. \\ \hline
\color{red} Original &  \color{red}sh*t is crazy around here \\ \hline
Detoxed & It is crazy around here. \\ 
& Stuff is crazy around here. \\
& Something is crazy around here.\\ \hline
\color{red} Original & \color{red} delete the page and shut up \\ \hline
Detoxed & Delete the page and stay silent. \\ 
& Please delete the page. \\
& Delete the page.\\ \hline
\color{red} Original & \color{red} massive and sustained public pressure is the only way to get these
b*stards to act. \\ \hline
Detoxed & Massive and sustained public pressure is the only way to get them to act. \\ 
& Massive and sustained pressure is the only way to get these people to act. \\ \hline
\color{red} Original & \color{red} f*ck you taking credit for some sh*t i wanted to do \\ \hline
Detoxed & You are taking credit for something I wanted to do \\ 
& You’re taking credit fro something i wanted to do. \\ \hline
\color{red} Original & \color{red} you gotta admit that was f*ckin hilarious though! \\ \hline
Detoxed & you got to admit that was very hilarious though! \\
& you gotta admit that was hilarious though! \\
\end{tabular}
\caption{Example sentences from ParaDetox parallel detoxification corpora. Sentence in red is original (toxic) sentence, below are its polite paraphrases. Note that for the purpose of an overall correctness explicit words are masked with "*".}
\label{tab:ParaDetox}
\end{table*}
\newpage
\subsection{Russian Dataset}\label{sec:appendix_2}

Table \ref{tab:RuDetox} shows examples from Russian parallel detoxification corpus. 

\begin{table*}[h!]
\centering
\begin{tabular}{l|l}
\hline
\color{red} Original & \color{red} \foreignlanguage{russian}{Х*рню всякую пишут,из-за этого лайка.v Долбо**изм.}\\ \textit{Translation} & \textit{They write all sorts of bullshit, because of this like. Stupidity.}\\ \hline
Detoxed &  \foreignlanguage{russian}{Чушь всякую пишут, из- за этого лайка.}\\
\textit{Translation} & \textit{They write all sorts of nonsense, because of this like.} \\ \hline
\color{red} Original & \color{red} \foreignlanguage{russian}{А нахрена тогда ты здесь это писал?}\\ \textit{Translation} & \textit{Why the f*ck did you post it here?}\\ \hline
Detoxed &  \foreignlanguage{russian}{Зачем ты это писал?}\\
\textit{Translation} & \textit{Why did you post it?} \\ \hline
\color{red} Original & \color{red} \foreignlanguage{russian}{Е*анутые. Отобрать оружие и лодку и штраф тысяч 500}\\ \textit{Translation} & \textit{F*ckers. Take away weapons, boat and give a fine of 500 thousand.}\\ \hline
Detoxed &  \foreignlanguage{russian}{Отобрать оружие и лодку и штраф тысяч 500.}\\
\textit{Translation} & \textit{Take away weapons, boat and give a fine of 500 thousand.} \\ \hline
\color{red} Original & \color{red} \foreignlanguage{russian}{Не поверишь, сколько е**нутых на планете.}\\ \textit{Translation} & \textit{You won't believe how many f*cked up people are on the planet.}\\ \hline
Detoxed &  \foreignlanguage{russian}{Не поверишь сколько таких на планете.}\\
\textit{Translation} & \textit{You won't believe how many people like that are there on the planet.} \\ \hline
\end{tabular}
\caption{Example sentences from Russian parallel detoxification corpora. Sentence in red is original (toxic) sentence, below are its polite paraphrases.}
\label{tab:RuDetox}
\end{table*}

\section{Generation Examples}\label{sec:appendix_b}

Table \ref{tab:results} contains detoxification examples for different models. It is notable that in some cases models generate almost the same results. This can be explained by the similarity of the training procedure and the fact that the reference answer was the same. 

\begin{table*}[h!]
\centering
\begin{tabular}{l|l}
\hline
\color{gray} Original & \color{gray} After all its hard to get a job if you're stupid.
\\ mBART $5000$ ENG & After all its hard to get a job if you don't understand.\\ 
mBART $3000$ ENG & After all its hard to get a job if you are not smart. \\ 
mT5 base (multilingual) & It's hard to get a job if you don't know. \\ \hline
\color{gray} Original &  \color{gray} Fuck minimum security, put him in real prison.\\
mBART $5000$ ENG & Put him in a real prison with minimum security. \\
mBART $3000$ ENG & I don't care about minimum security, put him in real prison. \\
mT5 base & I don't care about minimum security, put him in real prison. \\ \hline
\color{gray} Original & \color{gray} Sure he should get punished ,but three f*cking years?\\
mBART $5000$ ENG & Sure he should get punished, but three years?\\
mBART $3000$ ENG & He should get punished, but three years? \\
mT5 base & He'd get punished, but three years? \\ \hline
Paraphrase &  \foreignlanguage{russian}{Что у этих людей в башке!? походу насрато!}\\
\textit{Translation} & \textit{What's going on in these people's heads!? It looks like shit!} \\ \hline
mBART $5000$ ENG & \foreignlanguage{russian}{Что у этих людей в голове!? Походу бесчеловечно.}\\
mBART $3000$ ENG & \foreignlanguage{russian}{Что у этих людей в голове? Походу ненормально!} \\
mT5 base & \foreignlanguage{russian}{походу этих людей!? походу!} \\ \hline
\end{tabular}
\caption{Some detoxified sentences produced by our fine-tuned models. \textcolor{gray}{Gray text} refers to the original sentence, below are its paraphrases.}
\label{tab:results}
\end{table*}

\end{document}